%% file: main.tex
\pdfoutput=1
\documentclass[10pt,twocolumn,letterpaper]{article}

\usepackage[pagenumbers]{cvpr}

\usepackage{graphicx}
\usepackage{amsmath}
\usepackage{amssymb}
\usepackage{booktabs}
\usepackage{capt-of}
\usepackage{caption}
\usepackage{amssymb}
\usepackage{amsfonts} 
\usepackage{pifont}%
\usepackage{multirow}
\usepackage{tabularx}
\usepackage{balance}
\usepackage[export]{adjustbox}
\DeclareMathOperator*{\argmin}{argmin}

\newcommand{\xmark}{\ding{55}}%
\newcommand{\ours}{$\textit{R2O}$\xspace}
\usepackage[pagebackref,breaklinks,colorlinks]{hyperref}

\usepackage[capitalize]{cleveref}
\crefname{section}{Sec.}{Secs.}
\Crefname{section}{Section}{Sections}
\Crefname{table}{Table}{Tables}
\crefname{table}{Tab.}{Tabs.}

\def\cvprPaperID{4007}
\def\confName{CVPR}
\def\confYear{2023}

\begin{document}

\title{Refine and Represent: Region-to-Object Representation Learning}
\author{Akash Gokul\textsuperscript{1}\thanks{Denotes co-first authorship where co-first authors can prioritize their names in their resumes/websites.} , 
Konstantinos Kallidromitis\textsuperscript{2}\footnotemark[1] , 
Shufan Li\textsuperscript{1}\footnotemark[1] , \\
Yusuke Kato\textsuperscript{3}, 
Kazuki Kozuka\textsuperscript{3},
Trevor Darrell\textsuperscript{1}, 
Colorado J Reed\textsuperscript{1}
\\ \\ \textsuperscript{1}Berkeley AI Research \\
\textsuperscript{2}AI Lab, Panasonic R\&D Company of America \\
\textsuperscript{3}Digital and AI Technology Center, Technology Division, Panasonic Holdings Co. \\
{\tt\small \{akashgokul, jacklishufan\}@berkeley.edu, k.kallidromitis@us.panasonic.com}
}

\maketitle

\begin{abstract}
Object-centric self-supervised learning methods have recently led to state-of-the-art results on dense prediction tasks such as object detection and segmentation.
However, object-centric pretraining methods rely on fixed, off-the-shelf segmentation heuristics to identify objects in an image.
In this paper, we present Region-to-Object Representation Learning (\ours), which oscillates between predicting segmentation masks and using these masks to pretrain an encoding network.
\ours determines the segmentation masks by clustering encoded features.
\ours then pretrains an encoding network by performing region-to-region similarity learning, where the encoding network takes different views of an image and maps the segmented areas to similar encoded features.
\ours uses a \textit{region-to-object pretraining curriculum} which encourages the predicted segmentation mask to output a large number of regions early on and progressively fewer regions throughout training, which as we show, corresponds to a region-to-region and then object-to-object pretraining.
Representations learned using \ours lead to state-of-the art transfer performance for object detection and instance segmentation, and semantic segmentation. 
Additionally, \ours ImageNet pretrained models are able to surpass state-of-the-art in unsupervised object segmentation on the Caltech-UCSD Birds 200-2011 dataset without further training.
\end{abstract}

\input{1_introduction}

\input{2_related_works}

\input{3_method}
\input{4_results}
\input{5_conclusion}

\newpage

{\small
\bibliographystyle{ieee_fullname}
\bibliography{egbib}
}
\clearpage
\input{appendix}

\end{document}

%% file: 1_introduction.tex
\section{Introduction}
 
\begin{figure}[!t]
    \centering
    \includegraphics[width=0.9\linewidth]{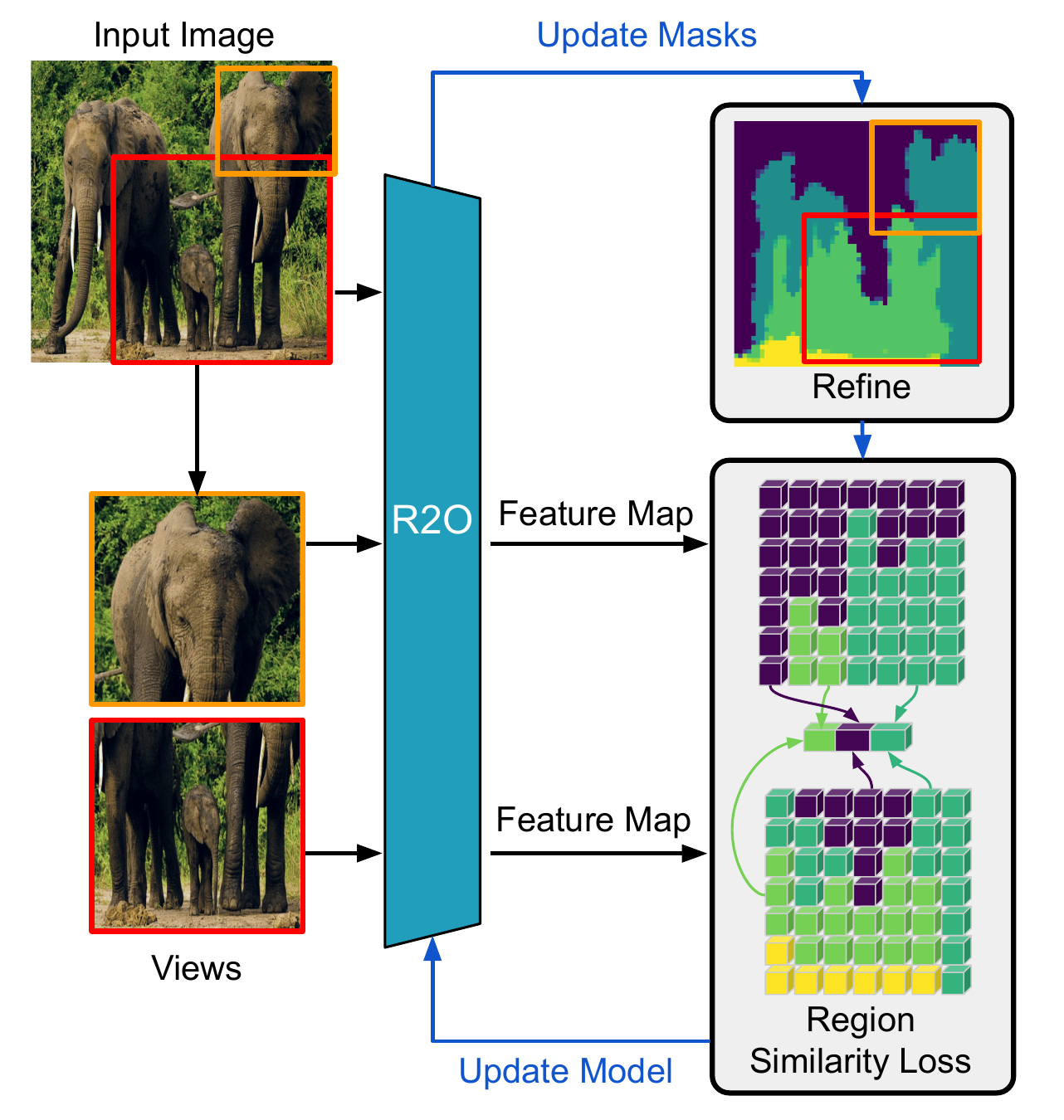}
    \caption{\textbf{\ours unifies region-based and object-centric self-supervised learning by jointly learning to discover and represent regions and objects.} Previous object-centric pretraining methods use off-the-shelf segmentation priors to learn object-centric representations. In contrast, \ours learns object-centric representations by osciallating between predicting a segmentation mask (that depends on the representations) and training representations (using the segmentation mask). Specifically, \ours pools features corresponding to segmented regions (region similarity loss) and enforces representation similarity for those features across views. Learned representations then lead to improved, increasingly object-centric masks, which in turn, lead to improved representations.
    }
    \label{fig:teaser}
\end{figure}

Progress in self-supervised pretraining in computer vision has led to improvements on a wide range of transfer learning tasks such as image classification, object detection, and semantic segmentation ~\cite{chen2020simple,he2020momentum,grill2020bootstrap,caron2021emerging,he2021masked}. Instance discrimination pretext objectives have driven many of these advancements, whereby contrasting augmented views of an input image, a network learns representations that are invariant to predefined image-level data augmentations. However, much of this work has focused on scene-level pretraining, which as recent works have shown, is not optimal for pixel-level prediction tasks, such as semantic segmentation ~\cite{wang2021dense, xie2021propagate, henaff2021efficient}. 

In order to train features which are better suited for dense prediction tasks, recent works have proposed extensions to the image-level pretraining objective. These approaches can be categorized into two types: region-based or object-centric. Region-based pretraining \cite{wang2021dense, o2020unsupervised, xu2021regioncl, xiao2021region} enforces predictive invariance between image patches, \ie regions, which are present in two views of an image. Thus, region-based methods can learn from all overlapping regions, regardless of size or content. While region-based methods outperform image-level pretraining methods on pixel-level prediction tasks, object-centric pretraining methods have led to better results~\cite{van2021unsupervised, henaff2021efficient, selvaraju2021casting, bar2021detreg, yang2021contrastive, xie2021unsupervised}. 

Object-centric pretraining uses segmentation heuristics to localize objects and enforces representational invariance to the features corresponding to those objects. In particular, DetCon \cite{henaff2021efficient} presents a simple, state-of-the-art approach which uses static object masks, generated by an off-the-shelf segmentation heuristic, in order to train object-centric features. However, because object-centric pretraining methods rely on off-the-shelf segmentation masks that are fixed throughout pretraining, their performance is limited by the accuracy of their object prior. In this paper, we seek an object-centric pretraining approach that does not rely on heuristics to localize objects but can instead discover object-centric regions while learning to represent them.

We present Region-To-Object representation learning (\ours): a framework that learns region-based and object-centric features by evolving predicted segmentation masks throughout pretraining and encouraging representational similarity for the features corresponding to the contents of the predicted mask. \ours uses a \textit{mask prediction module} to transform multiple small-scale image regions into larger object-centric regions using learned features. Over the course of pretraining, predicted masks are used to learn representations which in turn leads to more accurate, object-centric segmentations in future epochs (see \cref{fig:teaser}). This enables \ours to train object-centric features without relying on segmentation heuristics. 

Furthermore, we find that pretraining for dense prediction tasks benefits from both region-based and object-centric objectives at different stages of the pretraining process. Thus, we introduce a region-to-object curriculum, which gradually reduces the number of segments within predicted masks, in order to encourage region-level pretraining, corresponding to small image regions, early on and increasingly object-centric representation learning as pretraining progresses. This curriculum enables \ours to incorporate the strengths of region-based and object-centric pretraining as earlier epochs leverage a large number of image patches to train local features, as seen in region-based methods, and the later epochs train the object-centric features which are important for downstream tasks. 

We study \ours on a range of datasets and tasks, including object detection, instance segmentation, and semantic segmentation. Our contributions are summarized as follows: 
\begin{itemize}
\item We present \ours: a self-supervised pretraining method that learns region-based and object-centric representations by predicting a segmentation mask, using learned image features, and learning representations of the contents within the mask.

\item We introduce a region-to-object pretraining curriculum for \ours that begins by training local features corresponding to simple image regions, \eg adjacent pixels sharing similar color values, and gradually progresses to learn object-centric features. 

\item \ours pretraining improves transfer performance from ImageNet in object detection and instance segmentation on MS COCO 1$\times$ ($+0.2$ AP\textsuperscript{bb}, $+0.3$ AP\textsuperscript{mk}) and 2$\times$ ($+0.9$ AP\textsuperscript{bb}, $+0.4$ AP\textsuperscript{mk}), and semantic segmentation for PASCAL VOC ($+1.3$ mIOU) and Cityscapes ($+0.3$ mIOU) when compared to previous methods. After pretraining on COCO, \ours outperforms earlier methods by $+1.9$ mIOU for PASCAL VOC and $+2.4$ mIOU for Cityscapes. Additionally, \ours further improves state-of-the-art unsupervised segmentation performance on Caltech-UCSD Birds 200-2011 (CUB-200-2011) by $+3.3$ mIOU \textit{despite not finetuning on CUB-200-2011}.

\end{itemize}

%% file: 2_related_works.tex
\section{Related Works}
\paragraph{Image-Level Self-Supervised Pretraining}
Self-supervised learning aims to learn representations from unlabelled data in such a way that they are informative and effective when transferred to downstream tasks~\cite{bengio2013representation}. Early works in discriminative visual representation learning focused on training encoders to predict the position of an image patch ~\cite{doersch2015unsupervised, noroozi2016unsupervised}, predict the angle of rotation ~\cite{gidaris2018unsupervised}, and predict the colors in a grayscale image ~\cite{zhang2016colorful}. Current methods have shifted to training an encoder to learn representations that are invariant to data augmentations ~\cite{he2020momentum, chen2020simple, grill2020bootstrap, chen2021exploring, caron2021emerging, Reed_2021_CVPR, misra2020self}. This form of pretraining, known as instance discrimination, is often achieved through contrastive learning ~\cite{oord2018representation} or a teacher-student dual-network \cite{grill2020bootstrap}. The contrastive objective encourages representations from the same image to be invariant to data augmentations and dissimilar to representations from other images (commonly referred to as negatives). On the contrary, Siamese representational learning methods do not rely on negatives. ~\cite{grill2020bootstrap,chen2021exploring}.

\begin{figure*}[!t]
    \centering\includegraphics[width=0.8\textwidth]{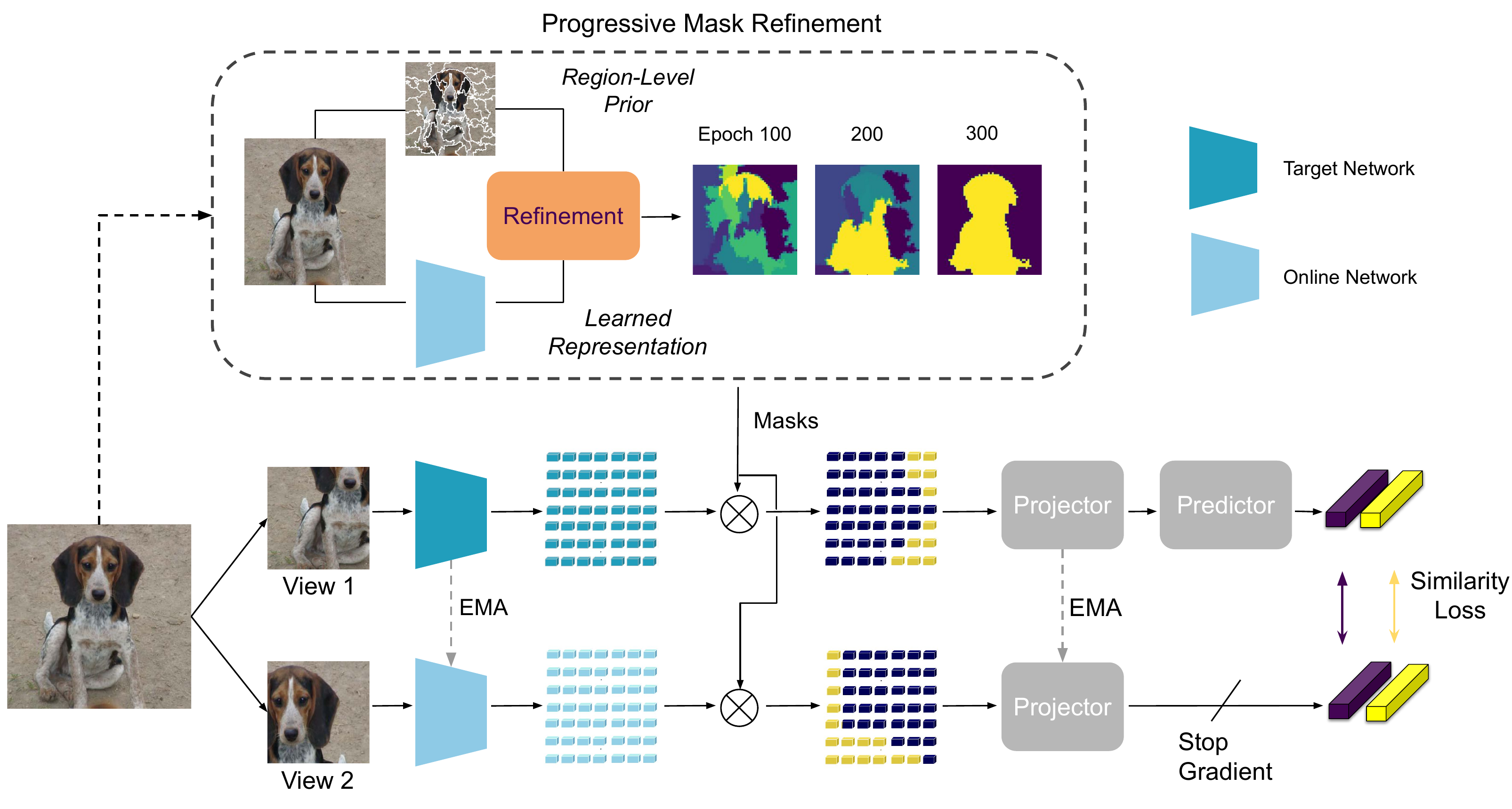}
    \caption{
    \looseness=-1 \textbf{The \ours architecture.} \ours consists of two interdependent steps: (1) \textit{Mask Prediction}, where small image regions (given by a region-level prior) are transformed into object-centric masks using learned features and (2) \textit{Representation Learning}, where object-level representations are learned. Step 1 produces object-centric segmentations by computing features for each region given by the region-level prior and performing K-means clustering on these region-level embeddings. Step 2 encourages representation similarity for the features corresponding to the contents of the mask in each view. Through this process, we have found that the features learned through the second step lead to increasingly object-centric segmentations which in turn train object-centric representations. Given the importance of region-level features in our mask prediction process, \ours employs a region-to-object curriculum which gradually decrease the number of clusters ($K$) during mask prediction from a very high value ($K=128$) to a low value ($K=4$), which as we show, enables region-based pretraining early on during training and slowly evolves to object-centric pertaining. We denote online and target network as ON, TN respectively.}  
    \label{fig:explainer}
\end{figure*}

\paragraph{Region-Based and Object-Centric Self-Supervised Pretraining}
Instance discrimination pretraining has primarily focused on learning image-level representations. While this is suitable for scene-level downstream tasks such as classification, it does not explicitly learn the locality of features which is critical for tasks such as object detection and semantic segmentation. Recent works ~\cite{wang2021dense, xie2021propagate, xiao2021region, wei2021aligning, o2020unsupervised,bar2021detreg, xu2021regioncl, xie2021detco} have extended the instance discrimination paradigm to also learn region-based features. These methods enforce representational invariance to image regions in different views without taking into consideration the content in these regions. Examples include DenseCL \cite{wang2021dense}, which trains local features to be invariant across views while also performing the standard image-level representation training, and ReSim \cite{xiao2021region}, which performs region-level representation learning by sliding a window across the image regions present in both views and enforcing region-based representational similarity. Our work extends region-based pretraining by first training region-level features and gradually discovering object-centric segmentations which eventually train object-centric features. 

Concurrent to this line of work are methods ~\cite{henaff2021efficient, wei2021aligning, bar2021detreg, yang2021contrastive, van2021unsupervised, selvaraju2021casting, xie2021unsupervised} that use heuristics ~\cite{felzenszwalb2004efficient,uijlings2013selective} to locate objects and enforce representational invariance for the features corresponding to these objects. The closest related work is DetCon \cite{henaff2021efficient}, which uses unsupervised segmentation algorithms to train object-level representations. In contrast, our method is able to discover and represent objects by evolving region-level segmentations into object-centric masks using learned features. Other works in this area ~\cite{wei2021aligning, van2021unsupervised, bar2021detreg,yang2021contrastive} focus on training object-centric features for a specific task, such as object detection, and do not train a network to learn features that can transfer to multiple tasks. Concurrent works such as Odin \cite{henaff2021efficient} and SlotCon \cite{wen2022self} have proposed object-level pretraining methods which do not rely on segmentation heuristics but instead use learned features to segment the input image. Similar to Odin, \ours uses K-means clustering on image features to help discover object-like regions. However, unlike Odin or SlotCon which focus on only training object-level features, \ours pretraining involves training both local features and object-level features.

\paragraph{Region Clustering}
\looseness=-1
In this work, we use clustering to refine small image regions into object-centric masks. Clustering has a long history in unsupervised semantic segmentation ~\cite{felzenszwalb2004efficient, achanta2012slic, dhanachandra2015image,hou2016dsets,comaniciu1999mean, comaniciu1997robust}. This trend has continued in the era of deep learning. Current methods utilize clustering assignments in order to learn spatially and semantically consistent embeddings ~\cite{kanezaki2018unsupervised, cho2021picie, hwang2019segsort, ji2019invariant}. For example, one popular approach \cite{kanezaki2018unsupervised, hwang2019segsort} involves trainining pixelwise embeddings by clustering these embeddings and using the cluster assignments as pseudo labels. Other works \cite{cho2021picie,  ji2019invariant} train embeddings using a similarity loss and task-specific data augmentations. Our work does not focus on the task of unsupervised semantic segmentation, but instead, focuses on using clustering to localize objects for learning transferable representations. Our method uses K-means clustering \cite{lloyd1982least} because of its simplicity and effectiveness.

%% file: 3_method.tex
\section{Method}
\label{sec:method}

We present \ours, a self-supervised pretraining method that enables region-level and object-centric representation learning. At a high level, \ours transforms region-level image priors into object-centric masks through a \textit{mask prediction module} and encourages representational invariance for the features corresponding to the contents of the discovered masks (see \cref{fig:explainer}). The training process uses a simple region-level prior, such as SLIC\cite{achanta2012slic}, to generate a fixed set of fine-grained image regions. During pretraining, features for each region are pooled and clustered using K-means to create segmentation masks. A high number of clusters, $K$, leads to masks that group small image regions together. Conversely, we have found that using a lower number of clusters leads to semantically meaningful segmentations which tend to correspond to objects. \ours then promotes representational similarity for the features corresponding to the predicted masks, by aligning said features across different views of an image. Finally, our region-to-object curriculum enables region-level and object-centric representation learning by gradually progressing masks from regional fragments to object-centric segmentations.

\paragraph{Formulation} We formulate the pretraining objective as a bilevel optimization problem. In this formulation, the upper-level objective transforms a region-based prior into an object-centric mask (\cref{sec:mask_refinement}) and the lower-level objective optimizes representational invariance for mask-level features  (\cref{sec:representation_learning}). Let $X$ refer to a given image and $R$ be the region-level prior for the image (\eg SLIC superpixels). The outer loop computes a cluster assignment, $M$, which is used to segment the image $X$, according to a mask-level objective ($\mathcal{L_{\text{mask}}}$) given the representation $f_\theta(X)$ and the region-level prior $R$ (\cref{eq:bilevel_outer}). The inner loop optimizes the parameters, $\theta$, of the encoder, $f_\theta$, following any representation learning objective ($\mathcal{L_{\text{repr}}}$) with respect to the segmentations created using $M$ (\cref{eq:bilevel_inner}).

\begin{align}
    \min_{M} \quad & \mathcal{L_{\text{mask}}}(M; f_{\theta^{*}}(X), R) \label{eq:bilevel_outer}\\
    \textrm{s.t.} \quad & \theta^{*} = \argmin_{\theta} \mathcal{L_{\text{repr}}}(f_\theta(X); M) \label{eq:bilevel_inner}
\end{align}

\subsection{Object-Centric Segmentation via Mask Prediction} \label{sec:mask_refinement}
Our \textit{Mask Prediction} step discovers object-centric segmentations by clustering region-level features for regions given by a region-level prior, such as SLIC superpixels \cite{achanta2012slic}. Specifically, we define an object-centric segmentation as a set of clusters, $M$, which satisfy criteria such as color consistency, spatial proximity, and representational similarity. Region-level priors provide segmentations which satisfy image-level criteria, such as color consistency, and can be derived solely from a given image. These region-level priors are fixed during training and add negligible overhead as they can be generated prior to pretraining. However, as illustrated in \Cref{fig:explainer}, these simple region-level priors do not accurately segment objects in the scene, but rather, group together small neighboring regions. In order to transform our region-level prior into an object-centric segmentation, we build upon recent works, which show that representations learned during self-supervised pretraining can encode object-level semantics \cite{zhang2021looking, xie2021unsupervised, van2020scan}, and incorporate learned features as part of our mask prediction step. 

We approximate the solution to the upper-level optimization problem (\cref{eq:bilevel_outer}) by performing K-Means clustering on the set of region-level image features. For each region from our region-level prior, we compute an embedding, $p_i$, using a mask-pooling operation, where each region has a binary mask that selects the corresponding features in the convolutional feature map. This generates a set of region-level embeddings $P = \{p_1, ..., p_{|R|}\}$ where $|R|$ is the number of distinct regions generated by the region-level prior $R$. Next, we apply K-Means clustering to the region-level embeddings in $P$ in order to compute clusters $M = \{M_1, ..., M_K\}$ (\cref{eq:kmeans}). Below, we denote $|M_i|$ as the number of points assigned to cluster $M_i$ and $\mu_i$ is the mean of the points in $M_i$. Given cluster assignments $M$, we produce an object-centric segmentation, $m$, by assigning each pixel in the representation $f_\theta(X) \in \mathbb{R}^{H \times W \times D}$ to its nearest cluster center (see Mask in \cref{fig:explainer}).

\begin{equation}
\label{eq:kmeans}
\mathcal{L_{\text{mask}}}(M; f_\theta(X), R) = \frac{1}{K}\sum_{k = 1}^K
 \frac{1}{|M_k|}\sum_{p \in M_k} ||p - \mu_k||^{2}\\
\end{equation}

\subsection{Object-Centric Representation Learning} \label{sec:representation_learning}
Our representation learning process follows the BYOL objective~\cite{grill2020bootstrap}. Since optimizing this inner objective to convergence can be expensive, we approximate $\theta^{*}$ by minimizing $\mathcal{L_{\text{repr}}}(f_\theta(X); M)$ for one step using mini-batch gradient descent.

\paragraph{Setup} Given an image $X$, we create two views:  $x_1 =  \mathcal{T}_{\text{aug}}(X)$, $x_2 =  \mathcal{T}_{\text{aug}}(X)$, where $\mathcal{T}_\text{aug}$ is any image augmentation policy. We use $m$ to obtain $m_1$, $m_2$ corresponding to views $x_1$, $x_2$. Further, we use a Siamese architecture with two similar networks: an online encoder $f_\theta$ (denoted by parameters $\theta$) and a target encoder $f_\xi$ (denoted by parameters $\xi$). The target network is updated using an exponential moving average of the online network's parameters.

\paragraph{Pretraining} We begin by computing $f_\theta(x_1)$, $f_\xi(x_2) \in \mathbb{R}^{H \times W \times D}$. Next, we replace the commonly used global-pooling with mask-pooling. This operation generates the average embeddings within the view-specific masks $m_1$ and $m_2$. We compute $h_{\theta, 1}$ and $h_{\xi, 2}$ by applying mask-pooling to $f_{\theta, 1}$ and $f_{\xi, 2}$. Following the Siamese architecture, we compute $z_{\theta, 1}$ and $z_{\xi, 2}$ by forward-passing $h_{\theta, 1}$, $h_{\xi, 2}$ through the online network's projector and the target network's projector respectively. Finally, we forward-pass $z_{\theta, 1}$ through the online network's predictor to obtain $q_{\theta}(z_{\theta, 1})$. We adopt the L2 loss according to the BYOL architecture (\cref{eq:loss_byol}). To make this loss symmetric for the view, we perform the same computation after swapping the two views to obtain $z_{\theta,2},z_{\xi,1}$. Our final loss is provided in \Cref{eq:loss_repr}.

\newcommand{\norm}[1]{\left\lVert #1 \right\rVert}
\begin{equation}\label{eq:loss_byol}
\mathcal{L}_{\text{BYOL}}(z_\theta, z_\xi) = 2 - 2 \cdot \frac{q_\theta(z_{\theta}) \cdot z_{\xi}}{\norm{q_\theta(z_{\theta})}_{2} \cdot \norm{z_{\xi}}_{2}}
\end{equation}

\begin{equation}\label{eq:loss_repr}
\mathcal{L}_{repr}=\mathcal{L}_{\text{BYOL}}(z_{\theta,1},z_{\xi,2})+\mathcal{L}_{\text{BYOL}}(z_{\theta,2},z_{\xi,1})
\end{equation}

\subsection{Region-to-Object Curriculum}
Empirically, we have found that varying the number of segments in the predicted masks, from a high number to a low number, results in improved downstream performance and more accurate masks. Thus, we introduce a region-to-object curriculum which allows for \ours pretraining to learn both region-based and object-centric features. Specifically, we control the number of clusters $K$ introduced in the mask prediction step (\cref{sec:mask_refinement}). We have found that a region-to-object curriculum where $K$ begins at a high value, \eg $K = 128$, and is reduced to a low value, \eg $K = 4$, over pretraining works best in terms of downstream performance. In the Appendix (\cref{fig:mask_vis_appendix}), we provide a visualization of the masks predicted during pretraining in order to illustrate how this curriculum allows for predicted masks to become increasingly object-centric. 

%% file: 4_results.tex
\section{Results}

\begin{table*}[t!]
    \centering
    \begin{tabular}{c c c c c c c c c}
    \toprule
    & \multicolumn{2}{c}{COCO 1$\times$} & 
    & \multicolumn{2}{c}{COCO 2$\times$} \\
    \cmidrule{2-3} \cmidrule{5-6}
      Method & 
      AP\textsuperscript{bb} &
      AP\textsuperscript{mk} & &
      AP\textsuperscript{bb} &
      AP\textsuperscript{mk} & &
      PASCAL VOC &
      Cityscapes \\
      \toprule
      Supervised & 38.9 & 35.4 & & 40.6 & 36.8 & & 72.4 & 74.7\\
      MoCo v2 \cite{chen2020improved} & 38.9 & 35.4 & & 40.9 & 37.0 & & 73.9 & 75.6 \\
      BYOL\textsuperscript{\textdagger} \cite{grill2020bootstrap} & 40.6 & 37.5 & & 42.0 & 38.7 & & 75.0 & 75.8\\
      DenseCL \cite{wang2021dense} & 40.3 & 36.4 & & 41.2 & 37.3 & & 73.8 & 76.1 \\
      ReSim \cite{xiao2021region} & 39.3 & 35.7 & & 41.1 & 37.1 & &74.3 & 75.5 \\
      PixPro \cite{xie2021propagate} & 41.4 & - & & - & - & & 74.2 & 75.9\\
      DetCon\textsubscript{\emph{B}}\textsuperscript{\textdagger} \cite{henaff2021efficient} & 41.5 & 38.0 & & 42.1 & 38.9 & &76.0 & 76.2\\
      DetCo \cite{xie2021detco} & 39.4 & 34.4 & & 41.4 & 35.8 & &74.3 & 74.9 \\
      SlotCon\textsuperscript{*} \cite{wen2022self} & 41.3 & 37.8 & & 42.9 & 39.3 & & 76.2 & 76.1\\
      \midrule
      \ours & \textbf{41.7} & \textbf{38.3} & & \textbf{43.0} & \textbf{39.3} & &\textbf{77.3} & \textbf{76.6}\\
      \bottomrule
      
    \end{tabular}
    \caption{\textbf{Performance on COCO object detection and instance segmentation, and semantic segmentation for PASCAL VOC and Cityscapes following ImageNet pretraining.} All methods pretrained a ResNet-50 backbone and finetune a Mask-RCNN (R50-FPN) for COCO and a FCN for PASCAL VOC and Cityscapes. \textsuperscript{*}: Denotes concurrent work. \textsuperscript{\textdagger}: Results from re-implementation.}
    \label{table:det_table}
\end{table*}

\label{sec:results}
We evaluate \ours via common transfer learning paradigms by pretraining using the images in the train set of the ImageNet ILSVRC-2012 dataset \cite{russakovsky2015imagenet}. We evaluate representations when finetuned for object detection and instance segmentation on MS COCO \cite{lin2014microsoft} and semantic segmentation on PASCAL VOC \cite{everingham2010pascal} and Cityscapes\cite{cordts2016cityscapes}. We also explore pretraining on scene-centric data and provide experiments which evaluate transfer performance after pretraining using MS COCO \texttt{train2017}. Additionally, we investigate performance on unsupervised object segmentation for Caltech-UCSD Birds 200-2011 \cite{wah2011caltech} when using our ImageNet pretrained models. Lastly, we perform several ablations exploring the various formulations and implementation specifics of \ours. Unless otherwise stated, results for existing methods are computed using pretrained weights officially released by the authors. The license, PII, and consent details of each dataset are in the respective papers. The code and models used in this work will be made publicly available. 

\paragraph{Architecture}
We use a ResNet-50 \cite{he2016deep} architecture for all encoders in order to enable fair comparison with prior works. In the Appendix, we explore using a Vision Transformer (ViT) encoder \cite{dosovitskiy2020image}. Following BYOL, all projector and predictor heads are 2 layer MLPs which use batch normalization \cite{batchnorm} after the hidden layer. Our mask prediction step uses the C4 output from our target network. For efficiency purposes, we perform K-means over the mini-batch on each GPU. Our predicted mask $m$ is of shape (14, 14). We use RoIAlign \cite{he2017mask} to align $m$ with views $x_1$ and $x_2$. Our region-to-object representation learning step applies mask-pooling to the C5 output of the ResNet-50 similar to the existing use of global pooling.

\paragraph{Optimization}
We follow the optimization details of BYOL (full details in Appendix). Specifically, we use a base learning rate of 0.3, which is scaled linearly to the product of the batch size and $K$, and a weight decay of $10^{-6}$. The learning rate is decayed by a cosine schedule after a warmup period which is one percent of the total pretraining time, \eg 300 epochs of pretraining uses 3 warmup epochs. Our target network's parameter $\tau$, used in the exponential moving average (EMA), is increased over training following a cosine schedule. We use the LARS optimizer with a global batch size of 4096 distributed over 128 NVIDIA V100 GPUs using PyTorch 1.10 \cite{NEURIPS2019_9015}.

\paragraph{Mask Prediction}
We precompute $100$ regions for each image using SLIC. We use a cosine scheduler that decreases the value of $K$, used in $K$-means, from 128 to 4 over the course of training. See Appendix for details.

\subsection{Transfer Learning Details}
\label{sec:transfer_details}

\paragraph{Object Detection and Instance Segmentation: MS COCO}
\looseness=-1
We finetune our encoder as the backbone of a Mask-RCNN (R50-FPN) \cite{he2017mask} using Detectron2 \cite{wu2019detectron2}. We follow the 1$\times$ (12 epochs) and 2$\times$ (24 epochs) training schedule, training on MS COCO's \texttt{train2017} dataset and evaluating on the \texttt{val2017} dataset. We report average precision for bounding box predictions (AP\textsuperscript{bb}) and mask predictions (AP\textsuperscript{mk}). Further details such as variance across multiple random seeds are reported in the Appendix. 

\paragraph{Semantic Segmentation: PASCAL VOC and Cityscapes}
We evaluate our semantic segmentation performance by finetuning the encoder as the backbone of a FCN \cite{long2015fully} using MMSegmentation \cite{mmseg2020}. For PASCAL VOC, we finetune using the \texttt{train\textunderscore{aug2012}} dataset for 45 epochs. For Cityscapes, we finetune using the \texttt{train\textunderscore{fine}} dataset for 160 epochs. Our evaluation pipeline follows the architecture and hyperparameters reported by the authors of MoCo \cite{he2020momentum}. Performance is measured by mean intersection-over-union (mIOU) on \texttt{val2012} and \texttt{val\textunderscore{fine}} respectively.

\subsection{ImageNet-1K Transfer}
\label{sec:summary}
We use the BYOL data augmentation policy to generate two views during pretraining which lasts for 300 epochs.

\paragraph{Object Detection and Instance Segmentation}
\Cref{table:det_table} shows our performance on COCO object detection and instance segmentation. \ours outperforms prior methods when using the 2$\times$ schedule ($+ 0.9$ AP\textsuperscript{bb}, $+ 0.4$ AP\textsuperscript{mk}) and is competitive with these methods when using the 1$\times$ schedule ($+ 0.2$ AP\textsuperscript{bb}, $+ 0.3$ AP\textsuperscript{mk}). When compared to concurrent work (SlotCon), \ours demonstrates an improvement of $+0.4$ AP\textsuperscript{bb} and $+0.5$ AP\textsuperscript{mk} when using the 1$\times$ schedule. We attribute our state-of-the-art performance on instance segmentation to the nature of our pretraining. \ours segments a scene by refining a large number of small image regions into a few larger regions that are semantically meaningful while learning representations corresponding to each of the discovered segments. Thus, segmentation-based tasks should benefit from this form of pretraining as representations learned during pretraining rely upon the boundaries of discovered masks instead of rectangular bounding boxes.

\paragraph{Semantic Segmentation}

As seen in \Cref{table:det_table}, we outperform previously existing methods on PASCAL VOC ($+1.3$ mIOU) and Cityscapes ($+0.3$ mIOU) semantic segmentation. Compared to concurrent work (SlotCon), \ours provides an improvement of $+1.1$ mIOU and $+0.5$ mIOU on PASCAL VOC and Cityscapes respectively. We believe our improvements in semantic segmentation demonstrate the benefits of pretraining region-level and object-centric representations. \ours improves transfer performance on this task as pixel-level features are learned while also training the object-centric representations needed to identify the objects in the scene. In particular, the benefits of \ours is demonstrated by our state-of-the-art performance on the PASCAL VOC semantic segmentation task, a pixel-level classification task which is object-focused as most of its classes are also present in ImageNet-1K.

\subsection{COCO Transfer}
We further explore using \ours when training on COCO, a scene-centric dataset. In comparison to the curated, object-centric benchmark of ImageNet pretraining, COCO images often contain multiple objects. Thus, scene-centric datasets such as COCO provide a challenging benchmark for object-centric pretraining as they test the effectiveness of object discovery methods and the generalizability of the pretraining method. For these reasons, pretraining on scene-centric data has recently become an important benchmark in self-supervised learning \cite{van2021revisiting, xie2021unsupervised, henaff2021efficient, wang2021dense}. We perform an additional experiment to evaluate the transfer performance after pretraining on COCO \texttt{train2017}. We use the same architecture, hyperparameters, and region-to-object schedule, \ie K = 128 → K = 4, as detailed in our ImageNet-1K experiments. We evaluate transfer performance for semantic segmentation on PASCAL VOC and Cityscapes following the details in \cref{sec:transfer_details}. As seen in \Cref{table:coco_pt_results}, \ours outperforms prior methods when pretrained on scene-centric data by $+1.9$ mIOU on PASCAL VOC and $+2.4$ mIOU on Cityscapes while being competitive with the concurrent work of SlotCon.  We have also found the quality of \ours predicted masks to be comparable to those generated during ImageNet pretraining when measuring average best overlap (ABO) \cite{uijlings2013selective}. Specifically, at the end of pretraining, \ours predicted segmentations obtain $0.46$ ABO when pretraining on COCO and $0.49$ ABO when pretraining on ImageNet.

\begin{table}[t]
    \centering
    \begin{tabular}{ c | c c }
    \toprule
      Method & 
      PASCAL VOC &
      Cityscapes\\
      \midrule
      MoCo v2 \cite{chen2020improved} & 69.2 & 73.8 \\
      DenseCL \cite{wang2021dense} & 68.5 & 74.0 \\
      DetCon\textsubscript{\emph{B}}\textsuperscript{\textdagger} \cite{henaff2021efficient} & 72.4 & 73.7 \\
      SlotCon\textsuperscript{*} \cite{wen2022self} & 74.3 & 76.0 \\
      \midrule
      \ours & \textbf{74.3} & \textbf{76.4}\\
      \bottomrule
    \end{tabular}
    \caption{\textbf{Performance on PASCAL VOC and Cityscapes semantic segmentation (mIOU) following COCO pretraining.} After pretraining on scene-centric data, \ours demonstrates state-of-the-art transfer performance for semantic segmentation on PASCAL VOC and Cityscapes. \textsuperscript{*}: Denotes concurrent work. \textsuperscript{\textdagger}: Results from re-implementation.}
    \label{table:coco_pt_results}
\end{table}

\begin{figure}[t!]
    \centering
    \includegraphics[width=\linewidth]{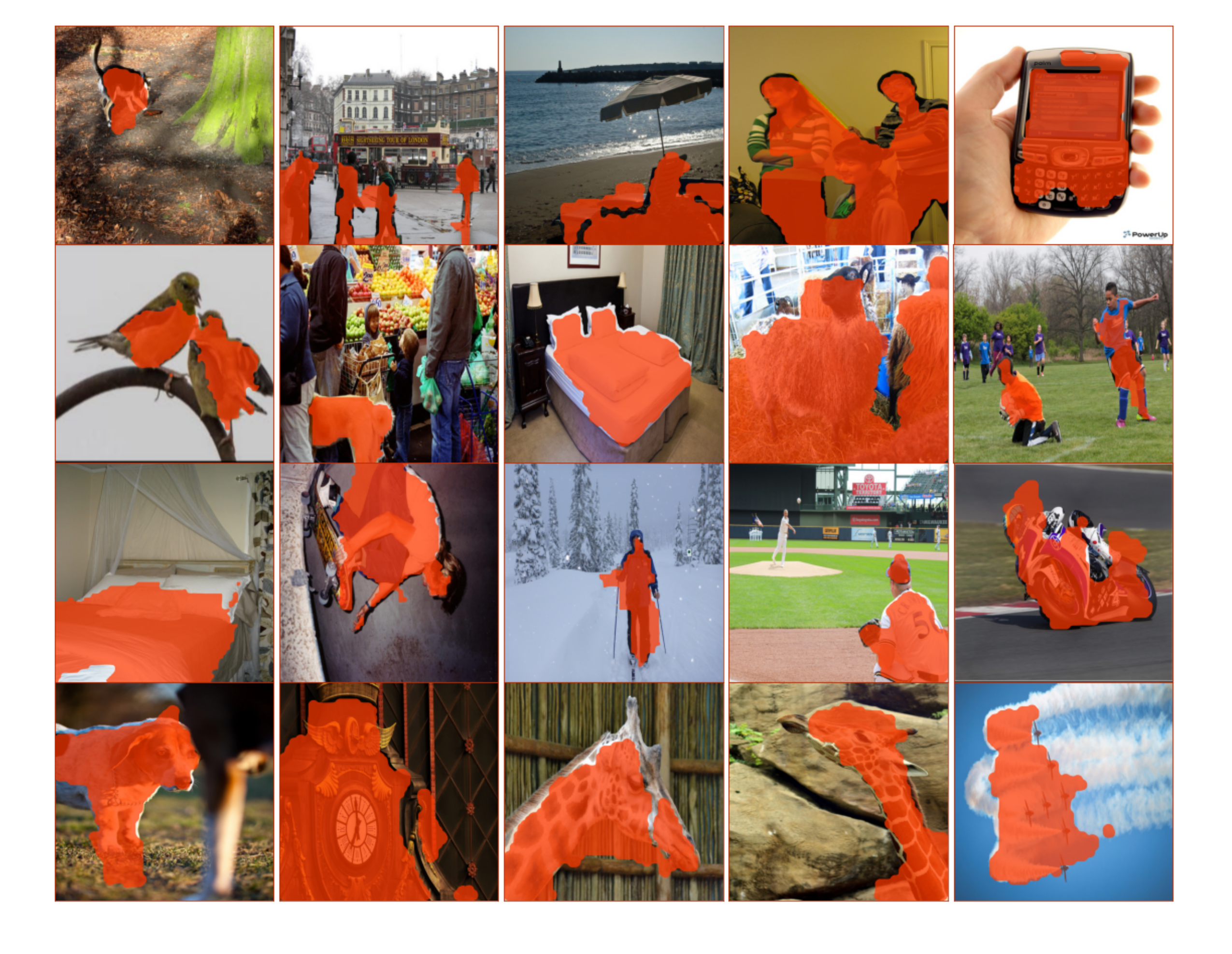}
    \caption{\textbf{Visualization of masks predictions during COCO pretraining}. In the later epochs of pretraining, \ours predicted masks are able to discover object-centric regions even in multi-object, scene-centric datasets such as COCO.}
    \label{fig:coco_pt_viz}
\end{figure}

\begin{table}[ht!]
    \centering
    \begin{tabular}{ c | c }
    \toprule
      Method & 
      CUB\\
      \midrule
    \textit{Self-Supervised Pretraining Methods} & \\
    MoCo v2 \cite{chen2020improved} & 63.5 \\
    BYOL\textsuperscript{\textdagger} \cite{grill2020bootstrap} & 65.1\\
    \midrule
    \textit{Unsupervised Object Segmentation Methods} & \\
    PerturbGAN \cite{bielski2019emergence} & 38.0 \\
    ReDO \cite{chen2019unsupervised} & 42.6 \\
    OneGAN \cite{benny2020onegan} & 55.5 \\
    Voynov \etal \cite{voynov2021object} & 68.3\\
    Melas \etal \cite{melas2021finding} & 66.4 \\
    \midrule
     \ours & \textbf{71.6} \\
     \bottomrule
    \end{tabular}
    \caption{\textbf{Performance on CUB-200-2011 segmentation.} Results report the mean intersection-over-union (mIOU) for foreground-background segmentations of images in the Caltech-UCSD Birds 200-2011 (CUB-200-2011) test set. Interestingly, we are able to outperform existing methods \cite{bielski2019emergence, chen2019unsupervised, benny2020onegan, voynov2021object}, \textit{without finetuning our ImageNet pretrained encoder}. \textsuperscript{\textdagger}: Results from re-implementation.}  
    \label{table:unsupervised_seg_table}
\end{table}

\subsection{Unsupervised Object Segmentation}
We investigate the quality of segmentations from \ours pretrained features by evaluating unsupervised object segmentation on Caltech-UCSD Birds 200-2011 (CUB-200-2011). We use our ImageNet pretrained ResNet-50 and generate object segmentations by applying K-means clustering to image features with $K=5$. We use Hungarian matching to assign one segment as the foreground and the remaining segments are considered as background. This approach accounts for the high variance in the background (\eg trees, ground, sky, etc.). \Cref{table:unsupervised_seg_table} shows our performance, measured by mean intersection-over-union (mIOU), when segmenting the foreground and background on a test set from CUB-200-2011 used in prior works \cite{chen2019unsupervised}. \Cref{table:unsupervised_seg_table} also provides baselines using self-supervised pretrained methods, such as MoCo v2 \cite{chen2020improved} and BYOL \cite{grill2020bootstrap}, following our evaluation protocol. In comparison to existing unsupervised object segmentation methods \cite{bielski2019emergence, chen2019unsupervised, benny2020onegan, voynov2021object}, segmentation via K-means clustering of \ours pretrained ResNet-50 features surpasses state-of-the-art ($+3.3$ mIOU) \textit{without finetuning on CUB-200-2011}. Similar to our protocol, Melas \etal \cite{melas2021finding} extract accurate segmentations from ImageNet pretrained GANs, without finetuning on CUB-200-2011, in order to train a UNet \cite{ronneberger2015u}. \ours  outperforms the work of Melas \etal ($+5.2$ mIOU) using only K-means clustering on features from \ours pretrained encoders. We provide visualizations of \ours segmentations in the Appendix.

\begin{table*}[ht!]
    \centering
    \begin{tabular}{c c c | c c}
        \toprule
        Mask Prediction & SLIC Prior & Region-to-Object Schedule & PASCAL VOC & ImageNet-100 \\
        \midrule
        \xmark & \xmark & \xmark & 66.4 & 69.0\\ 
        \xmark & \checkmark & \xmark & 67.0 & 69.0 \\ 
        \checkmark & \xmark & \xmark & 69.0 & 74.1 \\ 
        \checkmark & \xmark & \checkmark & 68.7 & 77.6  \\
        \checkmark & \checkmark & \xmark & 68.8 & 75.7 \\ 
        \checkmark & \checkmark & \checkmark & \textbf{72.1} & \textbf{85.5} \\ 
        \bottomrule
        \end{tabular}
    \caption{\textbf{Importance of \ours components}. We ablate the key components of \ours, namely: the mask prediction module, the use of a prior (SLIC), and the region-to-object schedule. We evaluate performance on PASCAL VOC semantic segmentation (mIOU) and $k$-NN classification (\%) on ImageNet-100. As shown, the combination of using Mask Prediction, SLIC Prior, and Region-to-Object Schedule is necessary for the top performing result (62.3 mIOU) -- removing any component affects performance by at least $-2.6$ mIOU and $-5.1\%$.}
    \label{tabel:r2o_ablation}
\end{table*}

\subsection{Ablation and Analysis}
\label{sec:ablation}
To examine the importance of the components within the \ours architecture, we perform ablation experiments to determine the importance of the mask prediction module, SLIC prior, and region-to-object curriculum. The models used in these ablations are pretrained on ImageNet-100 \cite{tian2019contrastive} for 300 epochs and evaluated on PASCAL VOC semantic segmentation and ImageNet-100 classification using $k$-nearest neighbors. We provide ablations on the effect of the type of scheduler used for our region-to-object curriculum (\eg cosine, linear), the type of pretraining curriculum (fixed $K$, object-to-region, region-to-object) and type of region-level prior (SLIC versus spatial segmentation) in the Appendix. The Appendix also explores using a contrastive representation learning objective.

\paragraph{Effect of Mask Prediction and Region-to-Object Pretraining.} \Cref{tabel:r2o_ablation} details the results of ablating each key component in \ours pretraining. The baseline method, which is a simple region-based pretraining method that does not use a SLIC prior, region-to-object schedule, nor the mask prediction module, performs the worst. The impact of the region-to-object curriculum on PASCAL VOC performance is similar to that of the SLIC prior ($+3.3$ mIOU versus $+3.4$ mIOU). However, the larger impact of the region-to-object curriculum on the $k$-NN evaluation ($+9.8\%$ versus $+7.9\%$) suggests that the region-to-object pretraining curriculum is the most important component, followed by the SLIC region-level prior. These findings can be justified by the architecture of our mask prediction module which relies on region-level features in order to discover larger object-centric regions. 

\paragraph{Comparison with Odin \cite{henaff2021efficient}.}
While we are unable to fairly compare with performance reported by Odin, due to their significantly longer pretraining schedule, we have re-implemented their method and pretrained Odin on ImageNet-100. \ours outperforms Odin by $+3.1$ mIOU for semantic segmentation on PASCAL VOC ($72.1$ mIOU versus $69.0$ mIOU) when both methods are pretrained on ImageNet-100. It should be noted that the authors of Odin did not provide publicly available code or weights.

\paragraph{Intuitions on R2O's effectiveness.}
In line with findings from previous work in object-centric pretraining, we have found that \ours learns to discover and represent semantically meaningful regions which tend to correspond to objects. \ours predicted masks segment object-centric regions but these segmentations may contain artifacts such as failing to distinguish different objects (see Appendix). These artifacts are also exhibited by off-the-shelf segmentation heuristics used in previously state-of-the-art object-centric pretraining methods, and thus highlight the effectiveness of learning to represent object-centric regions. \ours is able to find such regions by imposing a segmentation bottleneck in the form of the number of clusters, $K$, used in our mask prediction step. As pretraining enters the later epochs, the value of $K$ is low, forcing the mask prediction module to segment the input image into a few larger regions using the similarity between representations for various smaller regions in the image. The transfer performance of \ours after ImageNet pretraining and COCO pretraining demonstrate the effectiveness of this object discovery bottleneck, as it can handle both object-centric and scene-centric data.

\paragraph{Limitations.}\label{sec:limitations}
\looseness=-1
While our experiments and ablations indicate that \ours learns object-centric masks throughout pretraining, we cannot guarantee that such masks will necessarily identify objects. As shown by the examples in the Appendix, there are several failure cases where \ours segmentations group different objects together or separates parts of objects that have significantly different color, texture, or brightness compared to the rest of the object. \ours is an unsupervised pretraining algorithm, and without semantic grounding from object-level annotations, it cannot necessarily distinguish between the semantics of what constitutes an object. Nevertheless, as indicated by our strong experimental results, \ours learns useful representations that can be finetuned for dense downstream prediction tasks.

%% file: 5_conclusion.tex
\section{Conclusion}
\label{sec:conclusion}
We present Region-to-Object Representation Learning (\ours) which unifies region-based and object-centric pretraining by learning both region-level and object-centric representations using predicted masks which begin by clustering small-scale image regions and gradually progress to segment object-centric regions. While existing methods have either focused exclusively on region-based or object-centric pretraining, \ours shows that a region-to-object curriculum which transitions between these two objectives leads to state-of-the-art performance on a variety of downstream tasks. 

%% file: appendix.tex
\setcounter{section}{0}
\renewcommand{\thesection}{A.\arabic{section}}
\setcounter{equation}{0}
\renewcommand{\theequation}{A\arabic{equation}}
\section*{Appendix}

\section{Pretraining Details}
\label{sec:pretraining_details}
\paragraph{Data Augmentation} We follow the BYOL \cite{grill2020bootstrap} data augmentation pipeline. SLIC \cite{achanta2012slic} masks are generated using a resized (224 $\times$ 224) copy of the input image. To generate the augmented views, we implement the following operations using \texttt{torchvision} \cite{marcel2010torchvision}.

\begin{enumerate}
  \item Random Crop: Generate two distinct random crops from the original image. Afterwards we resize both views to $224\times224$.
  \item Horizontal Flip: We flip the views horizontally with a $0.5$ probability.
  \item Color Jitter: With an implementation probability of $0.8$ we set brightness equal to $0.4$, contrast = $0.4$, saturation = $0.2$ and hue = $0.1$.
  \item Grayscale: We convert the images from RGB to Grayscale with a $0.2$ probability.
  \item Gaussian Blur: We always apply blurring for the first view and with a $0.1$ probability for the second view. We use a $23\times23$ kernel with a standard deviation chosen uniformly at random between ranges $[0.1, 2.0]$.
  \item Solarize: Solarize only implemented on view 2 with a probability of $0.2$ using the PIL package and a threshold of $128$.
\end{enumerate}

\paragraph{Mask Prediction} We employ a cosine schedule for the number of segments, $K$, which starts at $K_0$ at epoch 0 and gradually decreases to $K_f$ between epoch $t_\alpha$ and $T_{epoch}$, where $T_{epoch}$ is the total number of epochs during the pretraining. \Cref{eq:cosine_schedule} expresses $K$ as a function of $t$, the epoch number. We choose $K_0=128,K_f=4$ in our experiments. We set $t_\alpha = 40$ during ImageNet-1K and ImageNet-100 pretraining and set $t_\alpha = 8$ during COCO pretraining.

\begin{equation}
\label{eq:cosine_schedule}
K(t) = 
\left\{
    \begin{array}{lr}
        K_0, & \text{if } t < t_\alpha \\
        K_f + cos (\frac{2(t-t_\alpha)} {(T_{epoch} - t_\alpha)\cdot \pi })  (K_0-K_f) & \text{otherwise } 
    \end{array}
\right\} 
\end{equation}

\paragraph{Optimization} We followed the optimization details of BYOL \cite{grill2020bootstrap}. The online network is updated by a LARS optimizer \cite{you2017scaling} with a base learning rate of 0.3 and and a weight decay of $10^{-6}$. The actual maximum learning rate during training is $\text{base\_lr} \times \frac{\text{global\_batch\_size}  }{256} \times K_0$. The learning rate increases linearly during the warmup period and then decreases following a cosine schedule. Throughout training, the learning rate is scaled by $K(t)$. In our experiment, the global batch size is 4096 and the warmup period is 3 epochs. The target network parameter $\xi$ is updated by $\xi^t \leftarrow (1-\tau^t) \cdot \theta^t + \tau^t \cdot \xi^{t-1}$ where $\theta$ is the online network parameter and $\tau$ increases from $0.99$ to $1.00$ following a cosine schedule.

\section{Transfer Learning Details}
\label{sec:transfer_details}
\paragraph{Object Detection and Instance Segmentation: MS COCO}
We adopt Mask-RCNN \cite{he2017mask} architecture with a ResNet-50 backbone and FPN. In the fine tuning stage, images are randomly flipped and resized to $u\cdot 1024$ on the longest side where $u$ is uniformly sampled in $[0.8,1.25]$. The images are then padded or cropped to $1024 \times 1024$. The aspect ratio is kept the same as original image. During the evaluation, the images are resized to $1024$ on the longest side. We finetune on COCO for the 1$\times$ (12 epochs) and 2$\times$ (24 epochs) schedule using stochastic gradient descent.  We swept between learning rates of $0.003,0.1,0.3$ for our model and our DetCon \cite{henaff2021efficient} implementation to ensure a fair comparison. We use a momentum of $0.9$, a weight decay of $4 \cdot 10^{-5}$, and a global batch size of $32$. The learning rate increases linearly in the first $1000$ epochs and drops twice by a factor of $10$ after $\frac{2}{3}$ and $\frac{8}{9}$ of the total training time. 

\paragraph{Semantic Segmentation: PASCAL VOC and Cityscapes} We adopt FCN \cite{long2015fully} architecture with a ResNet-50 backbone. We made the following architecture changes on the backbone according to MOCO \cite{he2020momentum}. The 3×3 convolutions in conv5 blocks have dilation 2 and stride 1. This
is followed by two extra 3×3 convolutions of 256 channels with BN and ReLU, and then a 1×1 convolution for per-pixel classification. We set dilation = 6 in the two extra 3×3 convolutions.All other implementation details are the default of MMSegmentation \cite{mmseg2020}, For PASCAL VOC, we fine tuned on train\_aug2012  and evaluated on val2012 dataset. The images are randomly flipped, scaled by a ratio randomly sampled in $[0.5,2.0]$, and then cropped to $513 \times 513$ during the training. The evaluation is performed on the original image size. We fine tuned for 30k iterations with a batch size of 16 and a weight decay of $10^{-4}$. The learning rate is dropped by a factor of 10 at $\frac{7}{10}$ and $\frac{9}{10}$ of the total training time. For Cityscapes, we fine tuned on train\_fine and evaluated on val\_fine dataset. The images are randomly flipped, resized to $2049 \times 1025$, scaled by a ratio randomly sampled in $[0.5,2.0]$, and then cropped to $769 \times 769$ during the training. The evaluation is performed on a resolution of $2049 \times 1025$. We fine tuned for 30k iterations with a batch size of 16, and a weight decay of $10^{-4}$. The learning rate is dropped by a factor of 10 at $\frac{7}{10}$ and $\frac{9}{10}$ of the total training time.To ensure a fair comparison, we followed the practice of \cite{goyal2019scaling} and evaluate each pretraining method under a variety of learning rates and report the best results. We also report the standard deviation across three random seeds. For PASCAL VOC, we swept between the learning rate 0.001, 0.003, 0.01, 0.03, 0.1. For Cityscapes, we swept between learning rate 0.004, 0.01, 0.04.

\section{Extended Experimental Results}
\label{sec:appendix_seed}
We conduct additional experiments and report the average performance on semantic segmentation and instance detection/segmentation with standard deviation across 3 different seeds in \Cref{table:seg_table_seed} and \Cref{table:det_table_seed}.
\begin{table}[ht]
    \centering
        \begin{tabular}{ c | c c }
        \toprule
          Method & 
          PASCAL VOC &
          Cityscapes\\
          \midrule
          Supervised & 72.4 & 74.7 \\
          MoCo v2 \cite{chen2020improved} & 73.9 $\pm$ 0.12 & 75.6 $\pm$ 0.09\\
          BYOL\textsuperscript{\textdagger} \cite{grill2020bootstrap} & 75.0 $\pm$ 0.22 & 75.8 $\pm$ 0.31\\
          DenseCL \cite{wang2021dense} & 73.8 $\pm$ 0.2 & 76.1 $\pm$ 0.12 \\
          DetCon\textsubscript{\emph{B}}\textsuperscript{\textdagger} \cite{henaff2021efficient} & 76.0 $\pm$ 0.14 & 76.2 $\pm$ 0.09\\
          ReSim \cite{xiao2021region} & 74.3  $\pm$ 0.28 & 75.5  $\pm$ 0.15 \\
          PixPro \cite{xie2021propagate} &  74.2 $\pm$ 0.44 & 75.9 $\pm$ 0.37 \\
          DetCo \cite{xie2021detco} & 74.3 $\pm$ 0.07 & 74.9 $\pm$ 0.46 \\
          SlotCon \cite{wen2022self} & 76.2 $\pm$ 0.24 & 76.1 $\pm$ 0.19 \\
          \midrule
          \ours & \textbf{77.2} $\pm$ 0.15 & \textbf{76.9} $\pm$ 0.38\\
          \bottomrule
        \end{tabular}
        \caption{\textbf{Performance on PASCAL VOC and Cityscapes semantic segmentation (mIOU) with standard deviation across 3 seeds.} \textsuperscript{\textdagger}: Results from re-implementation of pretraining method.}
        \label{table:seg_table_seed}
\end{table}

\begin{table}[ht]
    \centering
    \begin{tabular}{ c | c | c | c }
    \hline
      Method & 
      Epochs &
      AP\textsuperscript{bb} &
      AP\textsuperscript{mk}\\
      \hline
      BYOL\textsuperscript{\textdagger} & 300 & 40.3 ($\pm$ 0.10) & 37.5 ($\pm$ 0.08) \\
      DetCon\textsubscript{\emph{B}}\textsuperscript{\textdagger} & 300 &  41.5 ($\pm$ 0.03) &  38.0 ($\pm$ 0.09) \\
      \hline
      \ours & 300 & \textbf{41.7 ($\pm$ 0.02)} & \textbf{38.3 ($\pm$ 0.07)}\\
      \hline
    \end{tabular}
    \caption{\textbf{Performance on COCO object detection and instance segmentation with standard deviation following the 1$\times$ schedule averaged across 3 seeds.} \textsuperscript{\textdagger}: Results from re-implementation.}
    \label{table:det_table_seed}
\end{table}

\section{Exploring Contrastive Representation Learning} \label{sec:simclr_vs_byol}
We investigate using a contrastive objective function for $L_{repr}$. We follow the authors of SimCLR \cite{chen2020simple} and use the NT-Xent objective and employ the same architecture, data augmentation policy, and optimization details. We adopt the mask sampling strategy employed by DetCon \cite{henaff2021efficient} in order to reduce the memory overhead caused by the cross-GPU \texttt{gather} operation used in contrastive learning. \Cref{table:simclr_vs_byol} compares the performance when using the SimCLR objective or the BYOL objective. Both methods were pretrained on ImageNet-100 using the default region-to-object ($K = 128 \rightarrow K = 4$) schedule. The BYOL objective leads to an improvement of $+4.51$ on PASCAL and $+1.44$ on Cityscapes for the task of semantic segmentation. We note that the authors of DetCon \cite{henaff2021efficient} demonstrated similar results when using a contrastive objective, with DetCon\textsubscript{S} requiring up to 3 times the pretraining in order to match the performance of DetCon\textsubscript{B}.

\begin{table}[t]
    \centering
    \begin{tabular}{ c | c c }
    \toprule
        Objective & 
        PASCAL VOC &
        Cityscapes \\
      \midrule
      SimCLR NT-Xent & 62.99 $ \pm 0.02 $ & 69.99 $ \pm 0.17 $ \\
      BYOL & \textbf{67.5} $\pm 0.1$ & \textbf{71.43} $\pm 0.25 $\\
      \bottomrule
    \end{tabular}
    \caption{\textbf{Comparing contrastive and Siamese representation learning objectives.} We investigate using contrastive learning, via the Simclr NT-Xent objective, during \ours pretraining. We find that using a Siamese representation learning objective such as BYOL outperforms its contrastive counterpart when finetuned for PASCAL VOC ($+4.51$) and Cityscapes ($+1.44$) semantic segmentation. Results report mean intersection-over-union (mIOU).}
    \label{table:simclr_vs_byol}
\end{table}

\begin{table}[t]
    \centering
    \begin{tabular}{ c | c c }
    \toprule
        Method & 
        AP\textsuperscript{bb} &
        AP\textsuperscript{mk} \\ 
      \midrule
      MAE \cite{he2022masked} & 34.8 & 31.7\\
      \ours & \textbf{40.5} & \textbf{36.6} \\
      \bottomrule
    \end{tabular}
    \caption{\textbf{Examining COCO object detection and instance segmentation performance when pretraining a ViT}. We report Average Precision on bounding box (AP\textsuperscript{bb}) and mask (AP\textsuperscript{mk}) predictions for \texttt{val2017} after pretraining on ImageNet-100 for 300 epochs.}
    \label{table:vit}
\end{table}

\section{Exploring Vision Transformer Architectures}
\label{sec:vit}
Inspired by the recent progress in applications of Vision Transformer (ViT) architectures \cite{he2022masked, caron2021emerging, chen2021empirical, dosovitskiy2020image}, we experiment with applying \ours pretraining to a ViT-Base (Vit-B) architecture.We apply additonal transpose convolution layers during the pretraining on the feature maps in the final transformer block to get 7x7 and 14x14 feature maps to mimic the output of ResNet C5 and C4 output. We did not make any other changes to the architecture or the loss. We pretrained R2O and MAE for 300 epochs on ImageNet-100 with a global batch size of 1024. We used the exact same optimization set up as our ResNet experiments. The base learning rate is 0.3 following a cosine decay schedule. The momentum is 0.9 and weight decay is $10^{-6}$. For MAE, we strictly followed the optimization settings written in the paper \cite{he2022masked}. We evaluate the pretrained weights on MS COCO for the task of object detection and instance segmentation. We fined tuned for 1$\times$ schedule (12 epoch) using AdamW optimizer \cite{loshchilov2017decoupled}. The learning rate is 3e-4, the layer-wise decay rate is 0.75, and the drop path rate is 0.2. The learning rate is multiplied by 0.1 at epochs 9 and 11. \Cref{table:vit} compares MAE \cite{he2022masked} to \ours when using a ViT-B encoder after pretraining on ImageNet-100 for 300 epochs. The \ours pretrained encoder provides a $+5.7$ increase in AP\textsuperscript{bb} and a $+4.9$ increase in AP\textsuperscript{mk} relative to MAE. Thus, the benefits of \ours pretraining for dense prediction tasks also apply to newer architectures such as ViTs. 

\paragraph{Evaluating Alternate Curriculums.} \Cref{tab:varying_k_schedule} demonstrates that a region-to-object curriculum leads to the best transfer performance on PASCAL VOC ($+2.9$ mIOU) for semantic segmentation and for ImageNet-100 $k$-NN classification ($+5.5\%$). This schedule outperforms using a fixed $K$ and an opposite curriculum which increases $K$ from a low value to a high value. Using a fixed $K$ does not allow for learning both region based features and object centric representation learning. If the fixed value of $K$ is high, say $K = 64$, then refined masks will have many distinct regions and the representation learning step will train features corresponding to small image regions or object parts. A small fixed value of $K$, say $K = 4$, has the opposite effect as representations learned will simply match large image regions across views and cannot learn small region based features necessary for pixel level prediction tasks. The opposite curriculum, denoted as object-to-region, also leads to suboptimal performance. We believe this is because such a schedule does not allow for objects to be found and represented. Instead, early refined masks are just large image regions which lack semantic meaning as the network has undergone very little training. 

\begin{table}[t]
    \centering
    \begin{tabular}{ c | c c}
      \toprule
          \multicolumn{1}{c|}{Curriculum} & PASCAL VOC & ImageNet-100 \\ 
          \midrule
          \multirow{4}{*}{\rotatebox{90}{\usebox2}}
          16 & 68.7 & 76.3\\
          64 & 68.8 & 75.7\\
          160 & 69.2 & 73.9\\ 
          \midrule
          
         \multirow{3}{*}{\rotatebox{90}{\usebox1}}
           4 $\rightarrow$ 16 & 63.1 & 63.8\\
           4 $\rightarrow$ 64 & 69.1 & 75.1\\
           4 $\rightarrow$ 128 & 69.1 & 80.0 \\
          \midrule
          
          \multirow{3}{*}{\rotatebox{90}{\usebox0}}
           16 $\rightarrow$ 4 & 67.5 & 73.8\\
           64 $\rightarrow$ 4 & 71.8 & 84.4 \\
           128 $\rightarrow$ 4& \textbf{72.1} & \textbf{85.5}\\
      \bottomrule
    \end{tabular}
    \captionof{table}{\textbf{Comparing fixed $K$, object-to-region, and region-to-object curriculums.} We pretrain an encoder on ImageNet-100 while using alternate schedules for our mask prediction step. We report mean intersection-over-union (mIOU) on PASCAL VOC semantic segmentation and ImageNet-100 $k$-NN accuracy (\%). We refer to any $K$ schedule which starts from a low number and ends at a high number, \eg $K = 4 \rightarrow K = 128$, as object-to-region.}
    \label{tab:varying_k_schedule}
\end{table}

\section{Impact of Scheduler Type} \label{sec:scheduler_ablation}
We also experiment with the type of region-to-object scheduler. The cosine scheduler, which was used in our ImageNet-1K and COCO pretraining experiments, performs similarly to the linear scheduler on PASCAL VOC semantic segmentation ($72.1$ mIOU cosine scheduler versus $72.2$ mIOU linear scheduler) and ImageNet-100 $k$-NN classification ($85.5\%$ cosine scheduler versus $85.2\%$ linear scheduler) after pretraining on ImageNet-100. This result shows the benefit of region-to-object pretraining as similar performance can be achieved regardless of the type of scheduler used during pretraining.

\section{Importance of Region-Level Prior} 
\label{sec:neighborhood_prior_ablation}
We further compare using the SLIC region-level prior with a spatial heuristic which divides the image into 14$\times$14 square patches. We find that both priors lead to similar performance on PASCAL VOC segmentation ($72.1$ mIOU SLIC versus $72.7$ mIOU spatial), but using a SLIC prior leads to a larger, more significant difference ($+2.5\%$) in ImageNet-100 $k$-NN accuracy ($85.5\%$ SLIC versus $83.0\%$ spatial). This suggests that using a broader set of image-level criteria, \eg color resemblance and spatial proximity in the case of SLIC, to generate our region-level prior will improve downstream transfer performance.

\section{Quantifying the objectness of masks throughout pre-training.} 
\label{sec:abo_section}
\begin{figure}[t!]
    \centering
    \includegraphics[width=\linewidth]{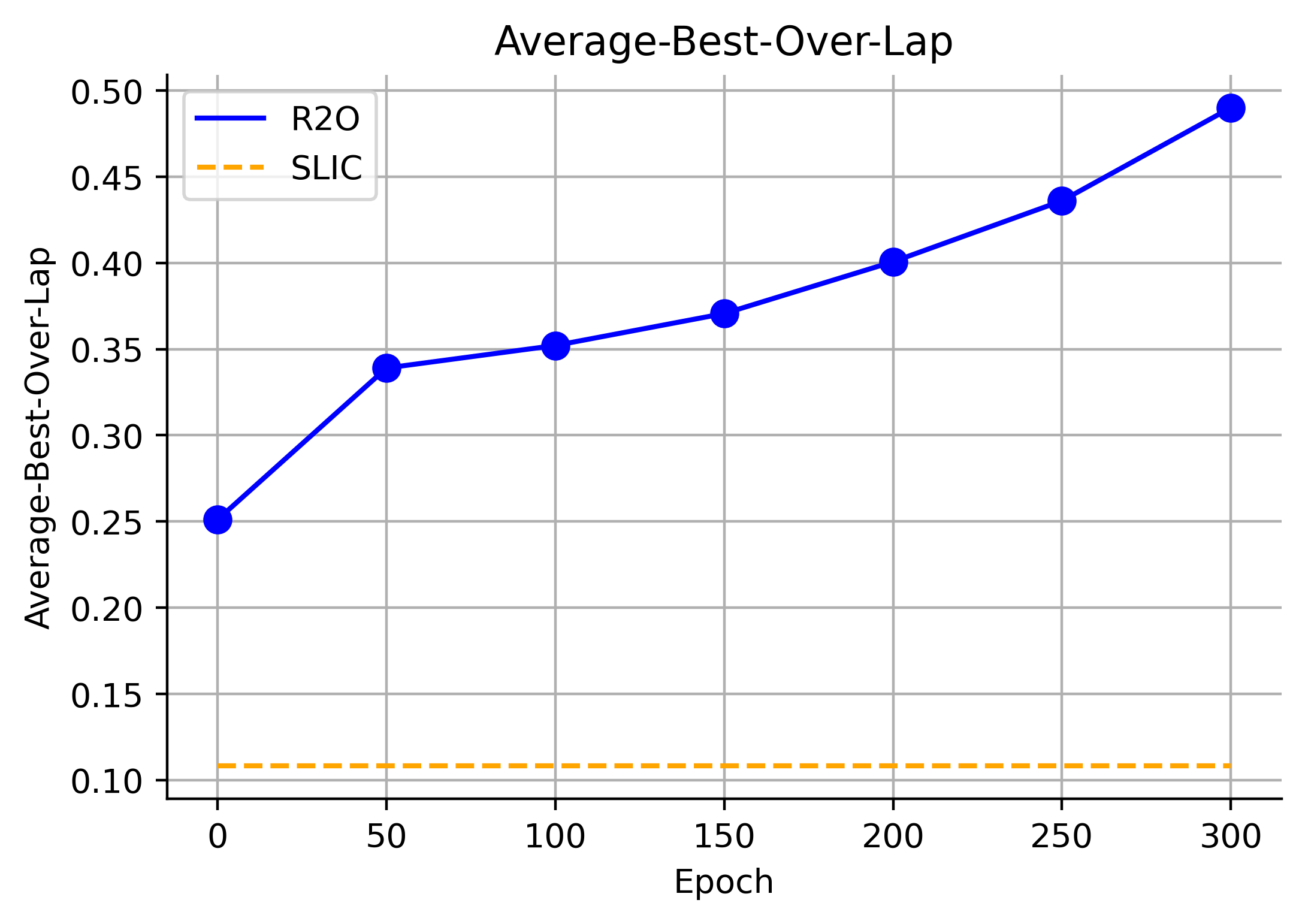}
    \caption{\textbf{\ours predicted masks become increasingly object-centric over pretraining.} We compute average best overlap (ABO) for ImageNet segmentations (using LUSS annotations) generated over the course of pretraining. For reference, we also display the ABO of our region-level prior (SLIC).} 
    \label{fig:ABO_graph}
\end{figure}

We use the Average Best Overlap (ABO)~\cite{uijlings2013selective} to measure the improvement of masks throughout training using semantic segmentation annotations for ImageNet-1K provided by LUSS \cite{gao2021large}. As shown in \Cref{fig:ABO_graph}, our ABO improved by $+0.25$ over the course of pretraining. \Cref{fig:ABO_graph} also includes the ABO of our SLIC prior as well. Masks predicted by \ours are consistently better than SLIC segments, suggesting that the mask prediction step is able to improve small region priors into more semantically meaningful object-like masks

\section{Odin Reimplementation Details}
We reproduced Odin \cite{henaff2022object} using a ResNet-50 encoder. We followed the optimization details mentioned by the authors. Specifically, the learning rate is linearly scaled by global batch size and decays following a cosine schedule. We trained Odin on ImageNet-100 for 300 epochs using a global batch size of 1024 across 32 NVIDIA V100 GPUs.

\section{Visualization of Mask Predictions}
\label{sec:visualize_mask}
\begin{figure*}[!ht]
    \centering
    \includegraphics[width=0.7\textwidth]{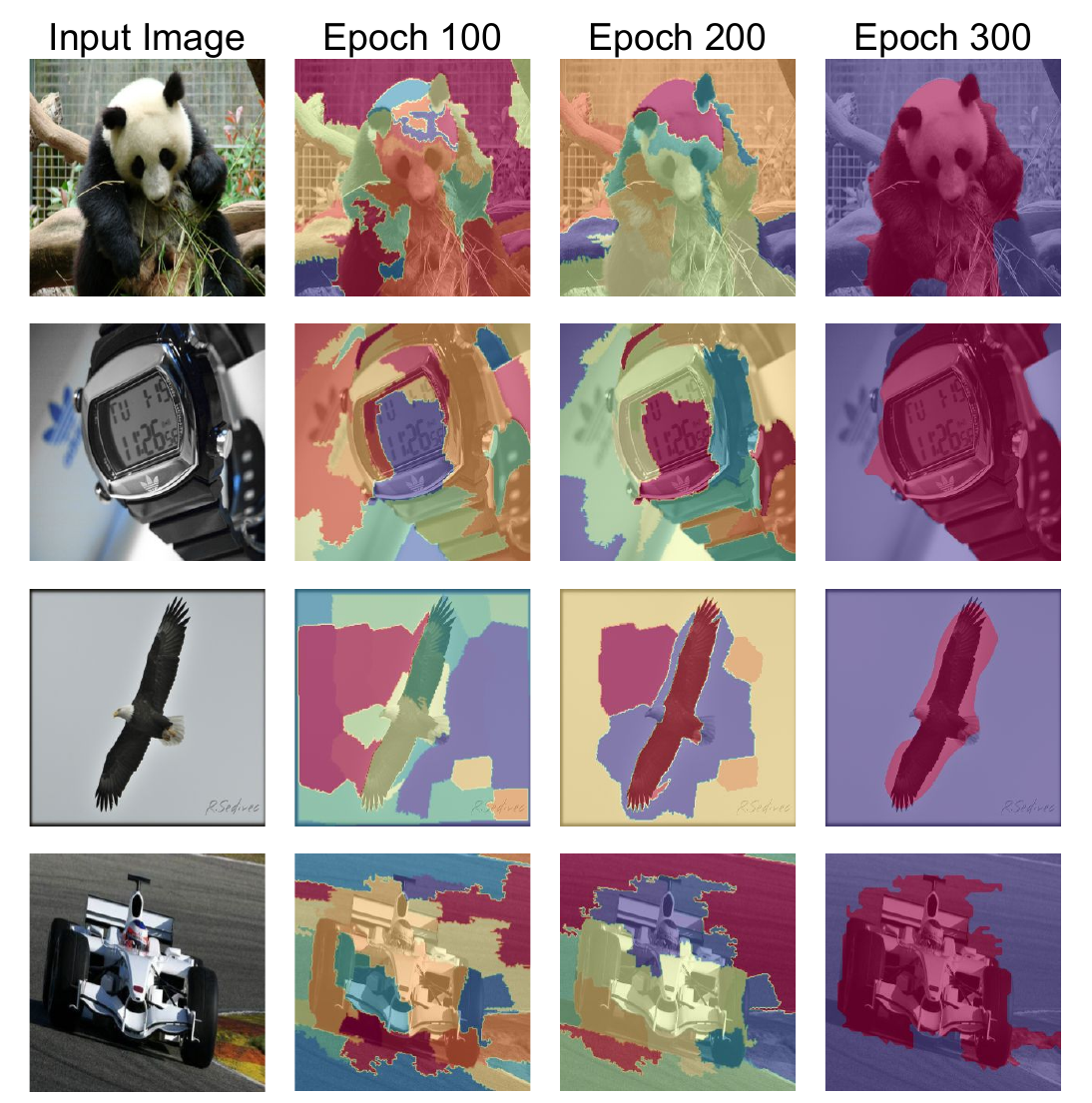}
    \caption{
    \looseness=-1 \textbf{Visualization of predicted masks generated during ImageNet pretraining} after 100, 200 and 300 epochs. As depicted, early masks consists of random image segments which gradually become object-centric segmentations.}
    \label{fig:mask_vis_appendix}
\end{figure*}

\paragraph{ImageNet-1K Pretraining}
In \Cref{fig:mask_vis_appendix}, we visualize the masks generated by \ours in the mask prediction step during pretraining. The number of segments, $K$, starts at 128 and gradually decreases to 4 following a cosine schedule. At 100 epoch, $K=119$. At 200 epoch, $K=74$. At 300 epoch, $K=4$. 

\begin{figure*}[!ht]
    \centering
    \includegraphics[width=0.7\textwidth]{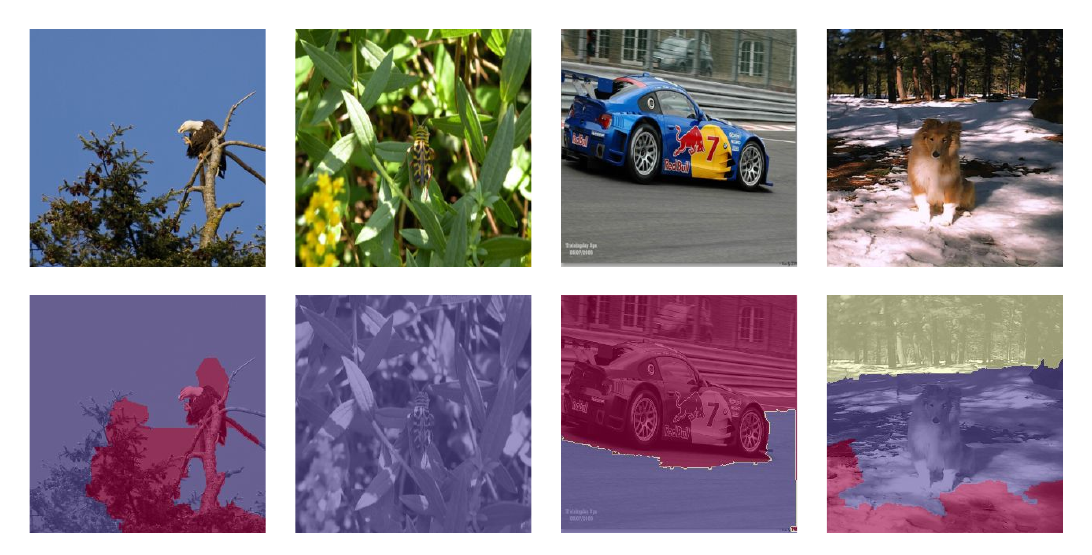}
    \caption{
    \looseness=-1  \textbf{Visualization of Failure Cases}. We visualize some of the poorly segmented masks generated during pretraining. From left to right: In example 1, one mask groups the eagle and part of the background together because of similar colors. In example 2, the mask is unable to detect any significant object in the image because the insect blends in the background. In examples 3 and 4, the masks focus more on the separation between the background and foreground and thus ignore the car and dog. }
    \label{fig:failure_cases}
\end{figure*}

In \Cref{fig:failure_cases}, we visualize some failure cases where the masks failed to capture objects. In these examples, masks group pixels that are similar in color and texture but do not belong to the same object (examples 1, 3). They also tend to over-segment regions into multiple parts if those parts have large disparities in color and texture such as the background, while at the same time, ignore smaller objects (example 2). In a few extreme cases, the masks groups all the pixels in entire image as belonging to the same object (example 2). Since we are performing K-means over the batch, this could happen when there is large variance across region-level features in the batch while the variance in region-level features from the given image is small. However, even in this worst case scenario, our representation learning step amounts to using the standard BYOL objective with global-pooling.

\section{Visualization of Caltech-UCSD Birds 200-2011 Segmentation}
\label{sec:visualize_cub_mask}
\begin{figure*}[!h]
    \centering
    \includegraphics[width=1\textwidth]{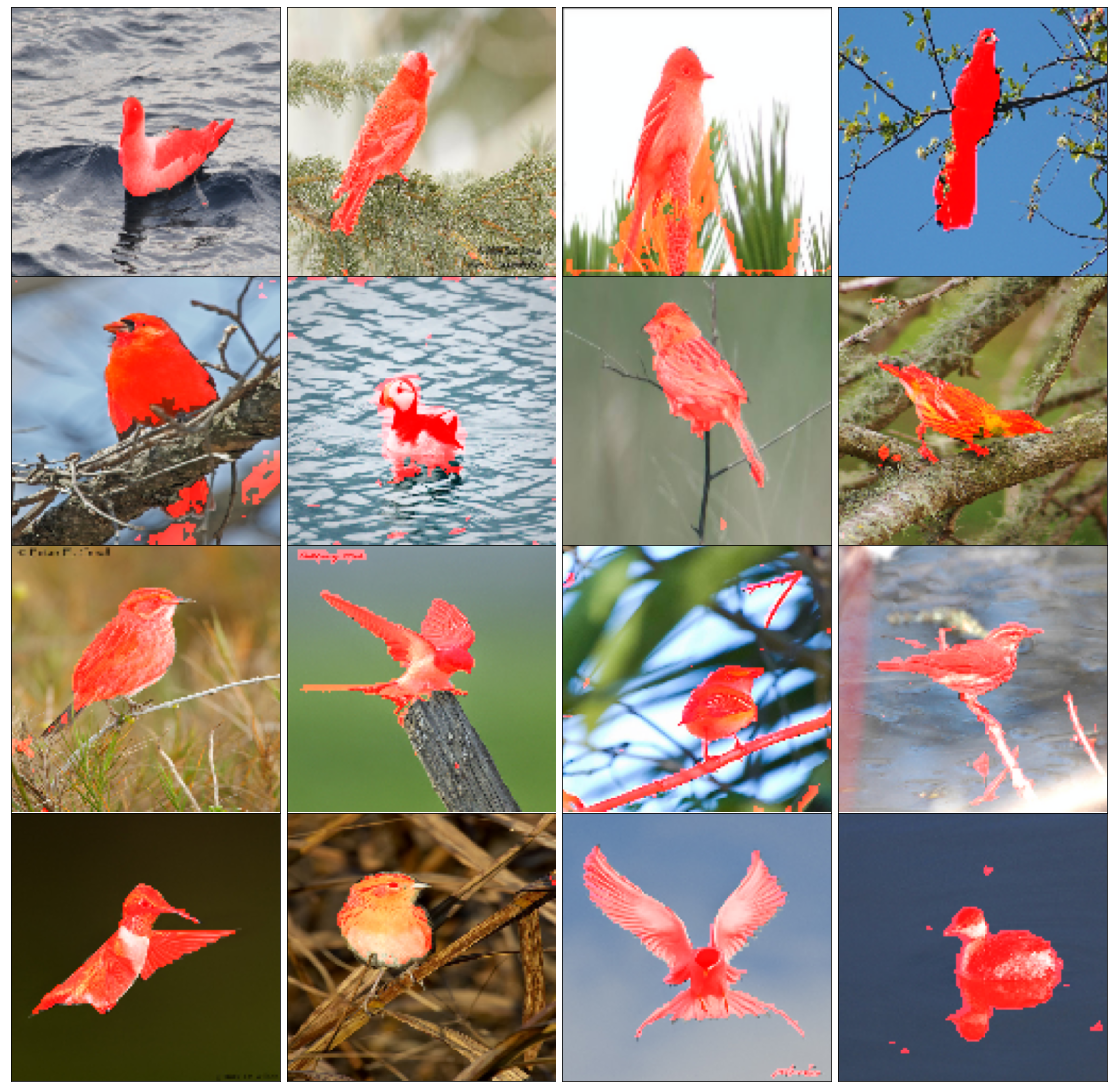}
    \caption{
    \looseness=-1 \textbf{Visualization of  Caltech-UCSD Birds 200-2011 Segmentation}. We perform unsupervised segmentation by clustering feature representations on a per-image basis with $K=5$. The best match is considered as the foreground.}
    \label{fig:vis_cub}
\end{figure*}

We also perform unsupervised segmentation on Caltech-UCSD with K-means clustering, setting $K = 5$, using \ours ImageNet pre-trained features. \Cref{fig:vis_cub} visualizes some of the masks.

